\RequirePackage[svgnames,table]{xcolor}

\documentclass[11pt,letterpaper,logo]{yalearxiv}

\usepackage{graphicx}
\usepackage{float,epstopdf}
\usepackage{bbm}

\usepackage{microtype}

\usepackage{natbib}
\setcitestyle{square}

\usepackage{subcaption}
\usepackage{booktabs}
\usepackage{amsmath}
\usepackage{amssymb}
\usepackage{mathtools}
\usepackage{amsthm}
\usepackage{dsfont}
\usepackage{multicol}
\usepackage{makecell}
\usepackage{multirow}
\usepackage{amsfonts}
\usepackage{mathrsfs}
\usepackage[amssymb, thickqspace]{SIunits}
\usepackage{enumitem}
\usepackage{pgfplotstable}
\usepackage{lipsum}		

\usepackage{microtype}
\usepackage{graphicx}
\usepackage{booktabs} 
\usepackage[table]{xcolor}
\usepackage{arydshln}
\usepackage[normalem]{ulem} 

\usepackage{cases}
\usepackage{wrapfig}

\usepackage{url}

\usepackage{thmtools}
\usepackage{thm-restate}
\usepackage{tabu}

\definecolor{huskypurple}{HTML}{4B2E83}

\usepackage{titletoc}
\usepackage{listings}
\lstdefinestyle{promptstyle}{
  basicstyle=\ttfamily\footnotesize,
  breaklines=true,
  breakautoindent=false,
  breakindent=0pt,
  postbreak=\mbox{\textcolor{gray}{$\hookrightarrow$}\space},
  columns=fullflexible,
  keepspaces=true,
  frame=single,
  framesep=5pt,
  xleftmargin=6pt,
  xrightmargin=6pt,
  aboveskip=8pt,
  belowskip=8pt,
  showstringspaces=false,
}

\newtcolorbox{simpleElegantQuote}{
    colback=AliceBlue!50!White,
    colframe=RoyalBlue!75!Black,
    boxrule=0.5pt,
    arc=2mm,
    boxsep=4pt,
    left=10pt, right=10pt,
    top=8pt, bottom=8pt,
    fontupper=\itshape,
}

\title{From Recursion to Convolution: Parallel Self-Improvement for Multi-Agent Systems}

\runningtitle{From Recursion to Convolution: Parallel Self-Improvement for Multi-Agent Systems}


\usepackage{fontawesome5}   
\usepackage{float}
\usepackage{graphicx}
\usepackage{caption}
\usepackage{adjustbox}
\usepackage{booktabs}
\usepackage{multirow}
\pgfplotsset{compat=1.18}
\definecolor{yaleblue}{RGB}{0,58,112}

\author{openJiuwen Team\\
Huawei Technologies Co., Ltd.\\
\faGithub~\textbf{Source Code:} \href{https://github.com/LYA2333/ConvRSI}{\texttt{https://github.com/LYA2333/ConvRSI}}
}

\hypersetup{colorlinks=true, linkcolor=blue!50!black, citecolor=blue!50!black,
            urlcolor=blue!50!black}

\begin{document}

\begin{abstract}
\vspace{-1mm}
\section*{Abstract} 
Multi-agent self-improvement enables agents to improve through interaction and feedback. However, sequential dependencies across agents and refinement rounds can propagate errors, increase computational costs, and limit parallelism. Moreover, refining individual features or agents in isolation can overlook their dependencies and restrict collective coordination. To address these challenges, we propose \texttt{ConvRSI}, a framework for parallelizing multi-agent self-improvement through convolution. \texttt{ConvRSI} couples intra-agent convolution, which jointly refines related agent features, with inter-agent graph convolution, which aggregates messages according to the collaboration topology. Both stages operate in latent space, enabling parallel updates without repeatedly decoding and re-encoding intermediate messages. Experiments on MATH-500 show that \texttt{ConvRSI} achieves 5.04\% performance improvement while reducing 17.27\% inference cost, and 45.11\% latency compared with RecursiveMAS. It also supports diverse collaboration patterns by representing their topologies as adjacency matrices. These results demonstrate the potential of convolutional co-evolution for efficient multi-agent self-improvement.

\end{abstract}

\maketitle

\section{Introduction}
\label{sec:intro}

Multi-agent systems (MAS) tackle complex tasks by coordinating agents with complementary roles, knowledge, and reasoning capabilities. Beyond collaboration among fixed agents, recent research explores how agents can improve through their interactions and experience. Prior work advances MAS through role-specific training feedback~\citep{motwani2024malt} and the co-evolution of question generation, solving, and judging through self-play~\citep{chen2026multi}. Recursive self-improvement (RSI) extends this direction by feeding improvements back into the system, enabling the updated system to guide subsequent rounds of improvement. For example, RecursiveMAS~\citep{recursivemas2026} iteratively refines shared latent states and jointly optimizes agents at the system level, while MASS~\citep{lee2026mass} alternates workflow evolution with model training on self-generated trajectories.
Despite this progress, scaling self-improvement to more agents and deeper iterations faces inherent challenges in optimization stability, computational efficiency, and the scope of collaboration, as discussed below.

\textbf{First, deeper recursion propagates error.} Each recursive round builds on earlier outputs or updates, allowing mistakes to influence subsequent decisions and become harder to correct. RecursiveMAS~\citep{recursivemas2026} jointly optimizes agents across rounds but retains the dependency chain through which errors can spread. Similar risks arise in other forms of agent RSI: inaccurate self-evaluation in MASS~\citep{lee2026mass} can introduce flawed trajectories into subsequent training, while mistaken memory updates in Recuris~\citep{yu2026recursive} can affect future task execution. This raises a key question: can comparable improvement be achieved with fewer sequential updates?

\textbf{Second, sequential multi-agent computation limits efficiency and parallelism.} Multi-agent RSI repeatedly executes agents, exchanges information, and evaluates updates, increasing computational costs as agent count and iteration depth increase. Chain-like dependencies force later updates to wait for earlier results, limiting the use of available parallel computing resources. Such dependencies arise across recursive collaboration links in RecursiveMAS~\citep{recursivemas2026}, the generation–verification–refinement pipeline in MALT~\citep{motwani2024malt}, and successive harness evolution cycles in ModularRSI~\citep{wu2026modularrsi}. This motivates restructuring agent updates to reduce the time and cost required to reach a target level of performance.

\textbf{Third, individual feature updates and chain-like evolution limit global coordination.} Each agent's behavior depends on related features, including task information, environmental observations, roles, experience, prompts, and messages. Refining one feature in isolation can overlook its dependencies on others. Existing methods address different aspects of improvement: Recuris~\citep{yu2026recursive} evolves memory, ModularRSI~\citep{wu2026modularrsi} evolves and integrates harness modules, MASS~\citep{lee2026mass} improves workflows with model capabilities, and RecursiveMAS~\citep{recursivemas2026} refines hidden states across agents. These leave room for a complementary design that jointly models related features within each agent and aggregates information across agents to support global evolution. In particular, chain-like evolution passes information from one agent to the next, making distant agents interact through intermediate updates. A broader approach would aggregate agents for collective, rather than agent-by-agent evolution~\citep{yu2026recursive, lee2026mass}.


To address these challenges, we propose \textbf{\texttt{ConvRSI}}, a framework for \emph{parallelizing multi-agent self-improvement through convolution}. The idea is to replace portions of sequential recursive computation with convolutional updates over agent features and multi-agent collaboration graphs. This introduces a \emph{space--time scaling perspective}: improvement can be pursued not only by increasing temporal iteration depth, but also by expanding spatial computation within each iteration. Rather than simply adding more independent agents, \texttt{ConvRSI} uses structured convolutional operations to coordinate parallel updates within and across agents.

\texttt{ConvRSI} comprises two coupled stages of improvement. \textbf{Intra-agent convolution} applies one-dimensional convolution to hidden-state sequences representing an agent’s task, role, experience, prompt, and messages. It jointly models local dependencies among these features, enabling each agent to independently refine its message state through the convolutional interactions. The features are \emph{initialized once} and subsequently refined through convolution. 
\textbf{Inter-agent convolution} then applies graph convolution to the updated agent messages according to the collaboration topology. Through parallel matrix operations, it aggregates information across connected agents. The resulting messages are fed into the next intra-agent stage, coupling individual refinement with collective evolution.
Both stages operate on latent representations, enabling efficient convolutional updates without repeatedly decoding messages into text and re-encoding them. This matrix-based design exposes parallelism across feature positions and agents throughout evolution. Moreover, \texttt{ConvRSI} supports diverse multi-agent collaboration patterns by representing their topologies as adjacency matrices for inter-agent graph convolution.

Empirically, we evaluate \texttt{ConvRSI} on 2 benchmarks spanning math and coding, using diverse model families and multi-agent collaboration patterns, including sequential style and mixture style. 
Compared with representative multi-agent and recursive baselines, \texttt{ConvRSI} improves task performance while reducing the cost of self-improvement. 
In particular, it achieves a 5.04\% performance improvement, a 17.27\% reduction in inference cost, and a 45.11\% inference speedup. At comparable performance, \texttt{ConvRSI} achieves these gains with $1.82\times$ and $1.80\times$ inference speedups and generates 17.3\% and 4.4\% fewer tokens per problem, demonstrating the efficiency of convolutional co-evolution.




Our main contributions are:
\begin{itemize}
    \item We introduce a \textbf{space--time perspective on multi-agent self-improvement}, exploring how spatially parallel computation can reduce reliance on deep sequential updates.

    \item We propose \texttt{ConvRSI}, which couples intra-agent convolution over related agent features with inter-agent graph convolution over collaboration topologies. Together, these stages support iterative individual refinement and collective evolution.

    \item We evaluate \texttt{ConvRSI} across diverse benchmarks, model families, and collaboration patterns, measuring performance gains under matched time and computational costs, as well as reductions in time and cost at comparable performance.
\end{itemize}
\section{Related Work}
\label{sec:related_work}

\noindent\textbf{Recursive Self-Improvement of Agents.} Recursive self-improvement enables agents to improve their capabilities through feedback and accumulated experience. Existing work explores different targets of improvement, including model behavior, agent workflows, harness components, and memory. Specifically, RecursiveMAS~\citep{recursivemas2026} extends recursive computation to multi-agent collaboration by iteratively refining shared latent states and jointly optimizing agents through an inner–outer learning procedure. MASS~\citep{lee2026mass} alternates multi-agent workflow optimization with model training on self-generated trajectories, using collective behavior as a source of self-supervision. MetaRSI~\citep{tan2026metarsi} further explores meta-level self-improvement by coordinating data, harness, and model updates and adapting the strategy that schedules these updates.
Beyond model-level improvement, ModularRSI~\citep{wu2026modularrsi} evolves agent harness modules using execution experience, Recuris~\citep{yu2026recursive} updates agent memory based on feedback, and Dream-RSI~\citep{zheng2026dream} reuses discovery histories to improve exploration policies. 
Building on RSI, \texttt{ConvRSI} reorganizes the improvement process through convolutional updates, coupling individual agent refinement with collective evolution while reducing reliance on sequential recursive steps.

\smallskip
\noindent\textbf{Multi-Agent Training and Co-Evolution.} Multi-agent training methods improve collective performance by encouraging agents to develop complementary capabilities and learn from their interactions. MALT~\citep{motwani2024malt} trains specialized agents for generation, verification, and refinement, while MAPoRL~\citep{park2025maporl} applies multi-agent reinforcement learning to improve collaboration among language models. Multi-Agent Evolve~\citep{chen2026multi} uses interacting roles to generate training data and feedback for model self-improvement. Related approaches explore the evolution of agent populations, as in EvoAgent~\citep{yuan2025evoagent}, or the automatic optimization of agentic workflows, as in AFlow~\citep{zhang2025aflow}. 
\texttt{ConvRSI} jointly refines features within each agent and uses a graph to model collaboration among agents, enabling collective co-evolution.

\smallskip
\noindent\textbf{Parallel and Efficient Multi-Agent Collaboration.} Multi-agent systems improve performance through collaboration, but additional agents increase compute costs, while sequential interaction rounds increase latency and limit parallelism.
Recent work~\citep{wunderlich2026multi} compares the accuracy–compute trade-offs of self-consistency, self-refinement, multi-agent debate, and agent aggregation. Parallel self-refinement~\citep{wang2026learning} generates multiple candidate solutions concurrently and synthesizes them into an improved response. Studies of multi-agent debate~\citep{smit2023should} further highlight the importance of comparison with strong alternatives under matched computational budgets. RecursiveMAS~\citep{recursivemas2026} improves the efficiency of recursive collaboration through latent-state communication.
Complementing existing approaches to efficient collaboration, \texttt{ConvRSI} uses matrix-based convolution to enable parallel updates across agent features and connected agents. Its latent-space operations avoid repeated text decoding and re-encoding during evolution.

\section{Preliminary}
\label{sec:v3-preliminary}

\subsection{Hidden States in Autoregressive Text Generation}
\label{sec:v3-hidden}

An autoregressive language model generates a response one token at a time. Given an input context \(x\) and the previously generated tokens \(y_{<t}\), a causal Transformer~\citep{Vaswani2017} produces a contextual hidden state \(\mathbf h_{t-1}\in\mathbb R^{H}\). An output head maps this state to the distribution of the next token:
\begin{equation}
  p_\theta(y_t\mid x,y_{<t})
  =\operatorname{softmax}\!
    \bigl(\mathbf W_{\mathrm{out}}\mathbf h_{t-1}
          +\mathbf b_{\mathrm{out}}\bigr)_{y_t}.
  \label{eq:v3-next-token}
\end{equation}
The hidden state is a continuous representation of the preceding context, whereas the output distribution is defined over discrete vocabulary tokens. During teacher forcing, the model conditions on a reference prefix \(y^\star_{<t}\) and causal masking allows all reference positions in a given sequence to be processed in one forward pass. During free generation, the model instead conditions on its own preceding outputs.

Hidden states also provide a communication medium for a multi-agent system. Rather than decoding every intermediate thought into text, an agent can transform its hidden states into continuous input embeddings for another agent. Because agents may have different hidden widths and embedding spaces, these representations require learned mappings before they can be consumed by the receiving model.

\subsection{Convolutional Computation}
\label{sec:v3-convolution-background}

A one-dimensional convolution applies a shared local filter along an ordered axis~\citep{LeCun1998}. For feature vectors \(\mathbf x_f\), its output at position \(f\) can be written as
\begin{equation}
  \mathbf z_f
   = \sigma\!\left(\sum_{\delta\in\mathcal K}
        \mathbf W_\delta\mathbf x_{f+\delta}+\mathbf b\right),
  \label{eq:v3-cnn-background}
\end{equation}
where \(\mathcal K\) denotes the filter offsets and \(\sigma\) is a nonlinearity. In our setting, the ordered axis contains different feature types for the agent; each position on this axis corresponds to a feature type.

A graph convolutional network (GCN) aggregates information in the neighborhoods of a graph~\citep{Kipf2017}. Let \(\mathbf A\in\mathbb R_{\ge 0}^{N\times N}\) describe communication between \(N\) agents, using the convention that \(A_{ij}>0\) means that agent \(i\) receives information from agent \(j\). We add a self-loop at each node and define
\begin{equation}
  \widetilde{\mathbf A}=\mathbf A+\mathbf I,\qquad
  \widetilde D_{ii}=\sum_j\widetilde A_{ij},\qquad
  \mathbf S=\widetilde{\mathbf D}^{-\frac{1}{2}}
              \widetilde{\mathbf A}
              \widetilde{\mathbf D}^{-\frac{1}{2}}.
  \label{eq:v3-normalization}
\end{equation}
Multiplying a matrix of agent representations by \(\mathbf S\) aggregates the representations of each receiver's senders. In our implementation, this normalization can also be applied to a directed adjacency. In that case, \(\mathbf S\) is not required to be symmetric.

\subsection{Convolutional Multi-Agent Co-Evolution}
\label{sec:v3-coevolution}

We define \textbf{Convolutional Multi-Agent Co-Evolution} as a sequence of coupled representation updates. At the \textbf{intra-agent convolution} level, each agent combines its context features into a local message using a convolution over feature types. At the \textbf{inter-agent convolution} level, a graph convolution updates the messages jointly along the communication graph. The updated messages supply continuous prefixes for the current round and message features for the next round.

RecursiveMAS~\citep{recursivemas2026} performs multi-round collaboration in latent space, using InnerLinks to form continuous inputs and OuterLinks to transfer states between heterogeneous agents. \texttt{ConvRSI} augments this collaboration with a graph-message recurrence. The preceding GCN output supplies the recurrent message feature, and the underlying MAS routes agent execution states through its latent links.

\section{Methodology}
\label{sec:v3-methodology}

\begin{figure}[tbp]
  \centering
  \includegraphics[
    width=0.85\linewidth,
    clip
  ]{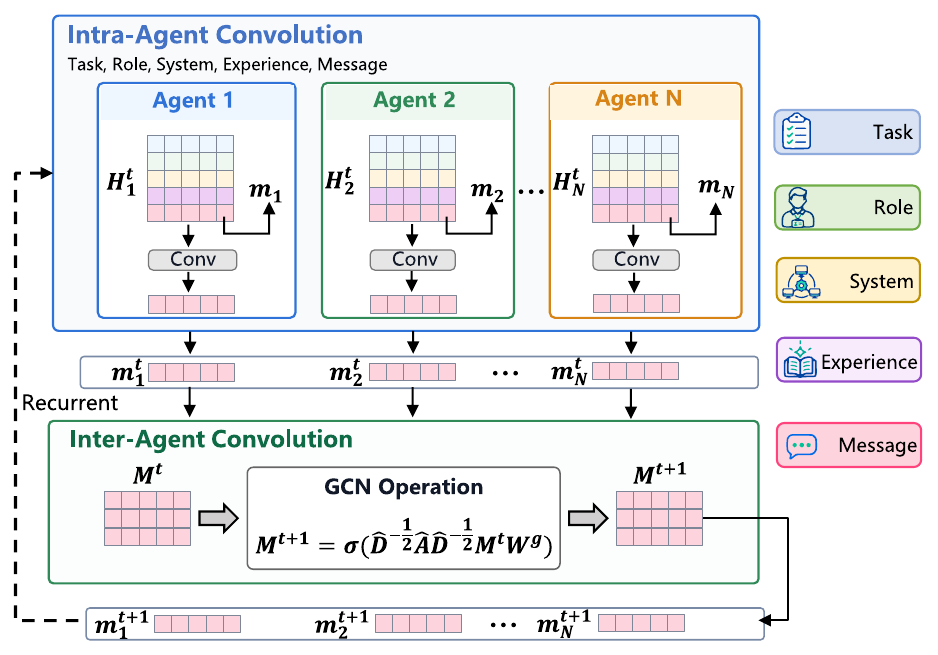}
  \caption{The \texttt{ConvRSI} framework shown for one slot across N agents. Intra-agent convolution produces a local message from each feature stack. The GCN aggregates the stacked local messages and returns one updated message per agent for the next round. The schematic uses \(M^t\) and \(M^{t+1}\) for the input and output of the graph layer, denoted by \(\mathbf U_k\) and \(\mathbf M_k\) in Eq.~\eqref{eq:v3-inter-conv}.}
  \label{fig:v3-framework}
\end{figure}

Consider a MAS with \(N\) agents and frozen language models \(\{\mathcal F_i\}_{i=1}^{N}\) of native hidden widths \(H_i\). The adjacency \(\mathbf A\) specifies which agents exchange convolutional messages. The MAS supplies an execution schedule \(\Pi\) for model invocations and latent transfers, as well as a final output interface. Figure~\ref{fig:v3-framework} shows a feature stack with \(F=5\) types ordered as \emph{task}, \emph{role}, \emph{system}, \emph{experience}, and \emph{message}. The first \(F-1\) are static text features, with experience optional. The message feature is recurrent. In the full-sequence setting, each feature has \(K\) slots. The convolutional operators use a shared width \(D\) and run for \(R\) rounds.


\subsection{Convolutional Co-Evolution}
\label{sec:v3-convolutional-coevolution}

\paragraph{Intra-Agent Convolution}
Let \(\mathbf Z_i\in\mathbb R^{F\times K\times H_i}\) contain the feature slots of agent \(i\). The first round masks the message feature and initializes its slots to zero. Subsequent rounds obtain it from the preceding GCN output. An agent-specific projection maps each feature slot to the common space as
\begin{equation}
  \mathbf x_{i,f,k}
    =\mathbf W_i^{\mathrm{in}}
       \operatorname{LN}\bigl(\mathbf z_{i,f,k}\bigr)
       +\mathbf b_i
    \in\mathbb R^{D}.
  \label{eq:v3-feature-projection}
\end{equation}
We apply the feature mask after projection and after adding learned feature-type embeddings. At each slot \(k\), we stack the resulting vectors in feature order to form the \(F\times D\) convolution input.

We then apply a channel-wise one-dimensional convolution along the feature axis. For a tensor \(\mathbf V\in\mathbb R^{F\times D}\), a convolutional block with the default residual connection is
\begin{equation}
  \mathbf V'
   =\operatorname{LN}\!\left[
       \mathbf V+
       \operatorname{GELU}\bigl(\operatorname{Conv}(\mathbf V)\bigr)
     \right].
  \label{eq:v3-intra-conv}
\end{equation}
The operator \(\operatorname{Conv}\) filters the feature axis independently in each channel, using a separate one-dimensional kernel per channel. The configured default has a kernel of size five, dilation one, and same padding. The convolution is applied independently to each slot and exchanges information across feature types while preserving the slot axis. In the optional no-residual variant, the block normalizes the activated convolution output without adding \(\mathbf V\); masks follow the actual convolution windows.

Let \(\mathbf v_{i,f,k}^{\mathrm{conv}}\) denote a feature vector after the final convolutional block. We apply a softmax to the gate scores \(g_i(\mathbf v_{i,f,k}^{\mathrm{conv}})\) over the valid positions \(\mathcal A_i\), yielding weights \(\beta_{i,f,k}\). A residual multilayer perceptron and a normalized projection \(\psi_i\) transform the pooled vector \(\overline{\mathbf v}_{i,k}=\sum_{f\in\mathcal A_i}\beta_{i,f,k}\mathbf v_{i,f,k}^{\mathrm{conv}}\) into a local message
\begin{equation}
  \mathbf u_{i,k}
    =\psi_i\!\left(\overline{\mathbf v}_{i,k}
           +\operatorname{MLP}_i(\overline{\mathbf v}_{i,k})\right).
  \label{eq:v3-local-message}
\end{equation}
We denote the full local operator, from feature projection to message readout, by \(\mathcal C_i\). Each agent has its own parameters and reuses them across rounds.

\paragraph{Inter-Agent Convolution}
For each slot, we stack the local messages into \(\mathbf U_k=[\mathbf u_{1,k};\ldots; \mathbf u_{N,k}]\in\mathbb R^{N\times D}\). One GCN layer updates all nodes using the same pre-update matrix:
\begin{equation}
  \mathbf M_k
  =\operatorname{LN}\!\left[
      \operatorname{GELU}\!\left(
          \mathbf S\mathbf U_k\mathbf W_g
          +\mathbf 1\mathbf b_g^{\mathsf T}
       \right)\right].
  \label{eq:v3-inter-conv}
\end{equation}
The \(i\)-th row, \(\mathbf m_{i,k}\), is the updated graph message for agent \(i\). The vector \(\mathbf 1\in\mathbb R^{N\times1}\) broadcasts the zero-initialized learnable bias \(\mathbf b_g\) to every agent. Layer normalization acts on the \(D\) channels of each row. The graph weights are shared among agents and rounds. Under Eq.~\eqref{eq:v3-normalization}, each agent aggregates the senders selected by \(\mathbf A\) and its own message via a self-loop. When slots are padded, aggregation uses the active nodes at that slot. The GCN updates those nodes synchronously.

\subsection{Parallel ConvRSI Training}
\label{sec:v3-training}

\begin{figure}[tbp]
  \centering
  \includegraphics[
    width=\linewidth,
    clip
  ]{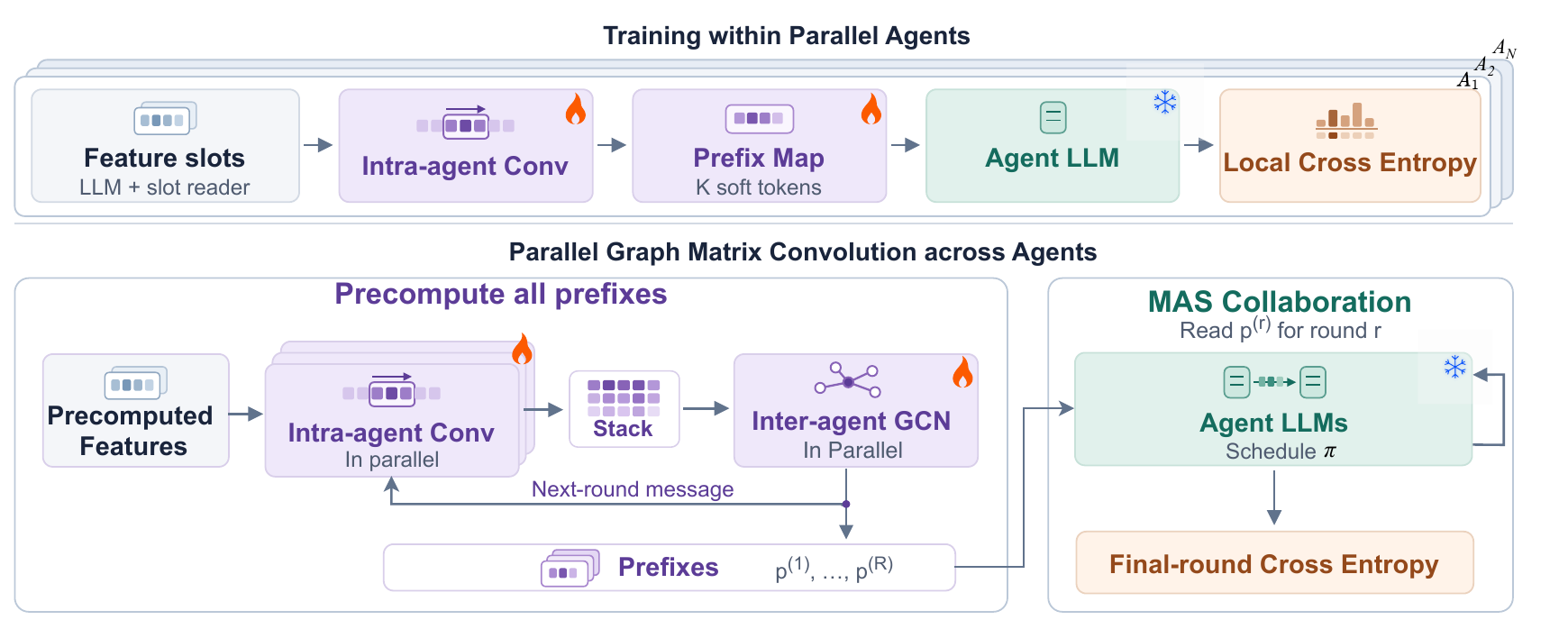}
  \caption{The \texttt{ConvRSI} training pipeline. Inner training uses independent agent objectives. Outer training can compute prefixes for all rounds through the convolutional recurrence before the collaboration forwards. The collaboration consumes the prefix for each round and computes cross-entropy at the final output. Flames highlight trainable components and snowflakes mark frozen language models.}
  \label{fig:v3-training}
\end{figure}

Figure~\ref{fig:v3-training} shows how the operators in Fig.~\ref{fig:v3-framework} are trained. Inner training learns to construct prefixes from each agent's local messages. Outer training jointly optimizes the graph-message recurrence and the latent transfer modules.


\paragraph{Training within Parallel Agents}
Each static component \(x_{i,f}\) is encoded independently by its agent's frozen language model. Let
\[
  \mathbf H_{i,f}
   =\operatorname{sg}\!\left[
      \mathcal F_i(\operatorname{Emb}_i(x_{i,f}))\right]
   \in\mathbb R^{L_{i,f}\times H_i},
\]
where \(\operatorname{sg}\) stops gradients through this static backbone computation. A trainable, agent-specific slot reader converts all valid hidden positions into \(K\) content-dependent slots:
\begin{align}
  \alpha_{i,f,\ell,k}
    &=\operatorname{softmax}_{\ell\in\mathcal T_{i,f}}\!\left(
        \bigl[\mathbf W_i^{\mathrm{st}}
          \operatorname{LN}(\mathbf h_{i,f,\ell})\bigr]_k
       \right), \nonumber\\
  \mathbf z_{i,f,k}
    &=\operatorname{LN}\!\left(
        \sum_{\ell\in\mathcal T_{i,f}}
          \alpha_{i,f,\ell,k}\mathbf h_{i,f,\ell}
        +\mathbf e_{i,k}^{\mathrm{st}}\right),
  \label{eq:v3-slot-reader}
\end{align}
where \(\mathcal T_{i,f}\) contains the non-padding positions. Each agent shares its slot reader across its static text components. We compute these slots once per example in a system forward pass and reuse them across the convolutional rounds.

In Inner training, each agent applies \(\mathcal C_i\) once with a masked message and maps the \(K\) resulting local messages to \(K\) soft prefix tokens. Its frozen LLM reads the prefix and predicts the local reference tokens under teacher forcing. The \(N\) losses in Eq.~\eqref{eq:v3-training-loss} are independent, so the agents can train concurrently.

\paragraph{Parallel Graph Matrix Convolution across Agents}
In Outer training, all agents can compute their local messages concurrently within a round. Stacking them yields a tensor of shape \([B,N,K,D]\) for a minibatch of \(B\) examples. We apply Eq.~\eqref{eq:v3-inter-conv} as a batched graph operation on the \(BK\) matrices of shape \(N\times D\). When the graph and node masks are fixed, the same normalization can be reused.

The superscript \((r)\) now indexes states and prefixes that vary by round. Let \(\mathcal Q_i:\mathbb R^D\to\mathbb R^{H_i}\) map a shared graph message to the input space of agent \(i\). It gives the recurrent feature at \(r>1\) and the prefix at every round as
\begin{equation}
  \mathbf z_{i,\mathrm{msg},k}^{(r)}
    =\mathcal Q_i(\mathbf m_{i,k}^{(r-1)}),
    \qquad
  \mathbf p_{i,k}^{(r)}
    =\mathcal Q_i(\mathbf m_{i,k}^{(r)}).
  \label{eq:v3-feedback-prefix}
\end{equation}
Let \(\mathbf Z_{\mathrm{static}}\) collect the static slots and \(\mathbf M^{(r)}\) collect the updated graph messages over all slots. We write \(\mathcal C\) for the local operators applied to their respective agents, \(\mathcal Q\) for the corresponding \(\mathcal Q_i\) applied to each graph row, and \(\mathcal G_{\mathbf A}\) for the slot-wise graph update in Eq.~\eqref{eq:v3-inter-conv}. For \(r>1\), the messages evolve as
\begin{equation}
  \mathbf M^{(r)}=\mathcal G_{\mathbf A}\!\left(\mathcal C\!\left(\mathbf Z_{\mathrm{static}},\mathcal Q(\mathbf M^{(r-1)})\right)\right).
  \label{eq:v3-graph-recurrence}
\end{equation}
The first round uses the masked zero message. The recurrence requires only the cached static slots and the previous GCN output. As shown in the lower left of Fig.~\ref{fig:v3-training}, we can run the convolutional rounds in order before the scheduled LLM forwards and store the resulting \(K\)-token prefix \(\mathbf p_i^{(r)}=(\mathbf p_{i,k}^{(r)})_{k=1}^{K}\) for each agent and round.

\paragraph{Learning Convolution as a Whole}
When \(\Pi\) invokes agent \(i\) in round \(r\), its LLM reads the prefix \(\mathbf p_i^{(r)}\) before its chat input and consumes the latent states supplied by the schedule. For a transfer from \(i\) to \(j\), the source InnerLink \(\mathcal I_i\) and the edge-specific OuterLink \(\mathcal O_{i\to j}\) map output hidden states to the receiver's input space, yielding \(\mathbf V_{i\to j}^{(r)}=\mathcal O_{i\to j}(\mathcal I_i(\mathbf H_{i,\mathrm{out}}^{(r)}))\). The schedule governs which states are read within a round and which are carried forward. Invocations with independent inputs can run concurrently. The latent links share parameters across rounds.

Each teacher-forced invocation obtains the reference-conditioned hidden states in one forward pass. A configurable token budget caps the length of each transferred sequence during training. The final output yields \(\mathcal L_{\mathrm{out}}\) in Eq.~\eqref{eq:v3-training-loss}. The stored prefixes retain their computation graph, so the final loss backpropagates through the graph-message recurrence, the frozen backbones, and the differentiable latent transfers.

\paragraph{Training Objectives}
The local and final token-level cross-entropy objectives are
\begin{equation}
  \mathcal L_{\mathrm{in},i}
    =\operatorname{CE}(\mathbf P_i,\mathbf y_i^\star),
  \qquad
  \mathcal L_{\mathrm{out}}
    =\operatorname{CE}(\mathbf P_{\mathrm{out}}^{(R)},\mathbf y^\star).
  \label{eq:v3-training-loss}
\end{equation}
Here \(\mathbf P_i\) and \(\mathbf P_{\mathrm{out}}^{(R)}\) are predicted vocabulary distributions at the local and final output positions. The corresponding targets are the reference token IDs \(\mathbf y_i^\star\) and \(\mathbf y^\star\). We average CE over supervised output tokens and then over examples. The Inner loss updates each agent's slot reader, local operator, and prefix map. The Outer loss jointly updates these modules, the GCN, and the OuterLinks, with frozen backbones and InnerLinks.

\paragraph{Collaboration Patterns}
The same convolutional operators can be applied to sequential, mixture, distillation, and deliberation styles by constructing a communication graph appropriate to the underlying MAS. These collaboration styles also appear in RecursiveMAS~\citep{recursivemas2026}. Table~\ref{tab:v3-patterns} gives representative adjacency constructions without fixing agent roles or backbone models. In every case, the degree matrix follows from \(\widetilde D_{ii}=1+\sum_j A_{ij}\); it is derived from the adjacency. The execution schedule, output interface, and any pattern-specific supervision are supplied by the MAS, while the intra/inter-convolution definitions remain the same.

\begin{table}[t]
  \centering
  \small
  \caption{Representative communication graphs for different collaboration
  styles. Rows of \(\mathbf A\) are receivers and columns are senders.
  Unspecified entries are zero. In the mixture construction, \(o\) is a
  designated receiving node, \(\mathcal B\) contains the remaining nodes,
  and \(w_j>0\). These are abstract graph constructions.}
  \label{tab:v3-patterns}
  \begin{tabular}{@{}p{0.16\linewidth}p{0.40\linewidth}p{0.35\linewidth}@{}}
    \toprule
    Style & Adjacency construction
      & Degree after adding self-loops \\
    \midrule
    Sequential & \(A_{i+1,i}=1\), \(1\le i<N\)
      & \(\widetilde D_{11}=1\); \(\widetilde D_{ii}=2\) for \(i>1\) \\
    \addlinespace
    Mixture & \(A_{o,j}=w_j\), \(j\in\mathcal B\)
      & \(\widetilde D_{oo}=1+\sum_{j\in\mathcal B}w_j\);
        \(\widetilde D_{jj}=1\) for \(j\in\mathcal B\) \\
    \addlinespace
    Distillation & Asymmetric influence:
      \(A_{ij}\ne A_{ji}\) on selected pairs
      & \(\widetilde D_{ii}=1+\sum_j A_{ij}\) \\
    \addlinespace
    Deliberation & Reciprocal exchange:
      \(A_{ij},A_{ji}>0\) on interacting pairs
      & \(\widetilde D_{ii}=1+\sum_j A_{ij}\) \\
    \bottomrule
  \end{tabular}
\end{table}

\paragraph{Inference Stage}
At inference, we encode the static inputs once, initialize the graph message with a masked zero, and use the same recurrence to supply prefixes to \(\Pi\). For each intermediate invocation, the InnerLink maps the last hidden state to the next continuous input. The OuterLinks route the resulting hidden states according to \(\Pi\). At the final output interface, the model autoregressively decodes the answer.

\subsection{Stability and Computational Complexity Analysis}
\label{sec:v3-analysis}

\paragraph{Stability}
For a fixed graph on the active agents, \(\mathbf S\) is similar to the row-stochastic matrix \(\widetilde{\mathbf D}^{-1}\widetilde{\mathbf A}\), so its spectral radius is one. A directed graph can yield \(\|\mathbf S\|_2>1\). For fixed static slots and feature masks after initialization, let \(L_{\mathcal C}\), \(L_{\mathcal Q}\), \(L_\sigma\), and \(L_{\mathrm{LN}}\) bound the Lipschitz constants of the local operator in its message input, the shared-to-native map, GELU, and LayerNorm. Two trajectories of Eq.~\eqref{eq:v3-graph-recurrence} satisfy
\begin{equation}
  \|\Delta\mathbf M^{(r)}\|_F \le \kappa\|\Delta\mathbf M^{(r-1)}\|_F,
  \qquad
  \kappa=L_{\mathrm{LN}}L_\sigma\|\mathbf S\|_2\|\mathbf W_g\|_2 L_{\mathcal C}L_{\mathcal Q}.
  \label{eq:v3-stability}
\end{equation}
Graph-message perturbations contract when \(\kappa<1\). For fixed teacher-forced references and latent sequence shapes under \(\Pi\), suppose the retained execution states \(\mathbf V^{(r)}\) satisfy
\begin{equation}
  \|\Delta\mathbf V^{(r)}\|
  \le a\|\Delta\mathbf M^{(r)}\|_F+b\|\Delta\mathbf V^{(r-1)}\|
  \label{eq:v3-system-stability}
\end{equation}
for uniform constants \(a,b\ge0\). Then \(\max(\kappa,b)<1\) is sufficient for both state perturbations to vanish over successive rounds.

\paragraph{Computational Complexity}
With \(J\) depthwise blocks of kernel width \(q\), the per-round forward costs of local message and prefix construction and dense graph propagation are
\begin{align}
  C_{\mathrm{local}} &= \mathcal O\!\left(BK\sum_{i=1}^{N}
    \left[FH_iD+JFqD+D^2+DH_i\right]\right), \label{eq:v3-local-complexity}\\
  C_{\mathrm{graph}} &= \mathcal O\!\left(BK\left[ND^2+N^2D\right]\right). \label{eq:v3-graph-complexity}
\end{align}
The local terms cover feature projection, depthwise convolution, pooled channel transforms, and mapping to the native space. The graph terms cover channel projection and dense aggregation. The \(R\) rounds cost \(R(C_{\mathrm{local}}+C_{\mathrm{graph}})\).

Let \(C_i(L;B)\) be the cost of an agent-\(i\) backbone forward at length \(L\). Static encoding costs \(\sum_i\sum_{f=1}^{F-1}C_i(L_{i,f};B)\) per system forward, and slot reading costs \(\mathcal O(BK\sum_{i,f}L_{i,f}H_i)\) over the same features. Teacher-forced collaboration adds \(\sum_a C_{i(a)}(L_a;B)\) over scheduled invocations \(a\) with agent \(i(a)\) and input length \(L_a\). At inference, \(Q_\Pi\) intermediate invocations of \(T\) latent steps each require \(Q_\Pi T\) latent forwards before final text decoding. Storing all round prefixes takes \(\mathcal O(BRK\sum_i H_i)\) entries, in addition to cached feature slots and training activations.

\section{Experiments}
\label{sec:experiments}

\subsection{Experimental Setup}
\label{sec:experimental_setup}




\paragraph{Datasets and metrics.} We use the Sequential-Math and Sequential-Code training datasets released with RecursiveMAS~\citep{recursivemas2026} for mathematical reasoning and code generation, respectively. We evaluate mathematical reasoning on MATH-500~\citep{lightman2024lets,math500} using final-answer accuracy and code generation on MBPP+~\citep{codeplus} using pass@1, the fraction of problems for which a single generated solution passes all benchmark tests. We report both metrics as percentages averaged over four random seeds.


\paragraph{Baselines and backbones.} We compare \texttt{ConvRSI} with five baselines. The first two use the Sequential-Light configuration of RecursiveMAS~\citep{recursivemas2026}: Recursive-TextMAS exchanges intermediate outputs as text, while RecursiveMAS passes latent states between agents. Both use Qwen3-1.7B as the planner, Llama-3.2-1B-Instruct as the critic, and Qwen2.5-Math-1.5B-Instruct as the solver. The remaining baselines are ModularRSI~\citep{wu2026modularrsi}, Hermes+~\citep{xue2026pastbench}, and MASS~\citep{lee2026mass}. We adapt all three to use the Sequential-Light solver backbone, Qwen2.5-Math-1.5B-Instruct~\citep{qwen2.5}, in place of the DeepSeek backbones used by ModularRSI and Hermes+ and the 27B Qwen backbone used by MASS.



\paragraph{Inference and evaluation.} We evaluate \texttt{ConvRSI}, Recursive-TextMAS, and RecursiveMAS at recursion depths $r\in\{1,2,3\}$. Following RecursiveMAS~\citep{recursivemas2026}, we use top-$p$ sampling with $p=0.95$ on both benchmarks. For MATH-500, we set the temperature to $0.6$ and the maximum generation length to 2{,}000 tokens. For MBPP+, we use a temperature of $0.2$ and a maximum of 4{,}000 tokens. Within each benchmark, all methods receive the same problem statement in the input format required by each method and use the same decoding settings and scoring procedure. We measure inference efficiency on NVIDIA RTX 4090D GPUs and report total runtime (seconds) and the mean number of generated tokens per problem.

\subsection{Main Results: Performance Across Recursive Depths}





We evaluate \texttt{ConvRSI} on MATH-500 and MBPP+ at recursion depths $r\in\{1,2,3\}$. Table~\ref{tab:main_results} reports MATH-500 accuracy, MBPP+ pass@1, total inference time on NVIDIA RTX 4090D GPUs, and generated-token counts. For the adapted ModularRSI, Hermes+, and MASS baselines, we use Qwen2.5-Math-1.5B-Instruct~\citep{qwen2.5}, the solver backbone in the Sequential-Light configuration. It replaces the DeepSeek backbones in ModularRSI and Hermes+ and the 27B Qwen backbone in MASS.

\paragraph{Performance across recursion depths.} On MATH-500, \texttt{ConvRSI} improves from 71.6\% accuracy at $r=1$ to 72.4\% at $r=2$ and 75.0\% at $r=3$. Its lead over the better of Recursive-TextMAS and RecursiveMAS grows from 0.2 to 1.4 to 3.6 percentage points across the three depths. Over the same range, RecursiveMAS improves from 70.1\% to 71.4\%, while Recursive-TextMAS declines from 71.4\% to 69.6\%. On MBPP+, Recursive-TextMAS leads at $r=1$ with 30.4\% pass@1, 0.4 points above \texttt{ConvRSI}. \texttt{ConvRSI} overtakes both recursive baselines at $r=2$ and reaches 32.9\% at $r=3$, 0.8 points above RecursiveMAS and 3.8 points above Recursive-TextMAS.

\paragraph{Inference efficiency.} \texttt{ConvRSI} completes both benchmarks in less time and generates fewer tokens than either recursive baseline at every depth. At $r=3$, it completes MATH-500 in 1{,}459\,s, compared with 2{,}658\,s for RecursiveMAS and 4{,}721\,s for Recursive-TextMAS, yielding $1.82\times$ and $3.24\times$ speedups, respectively. On MBPP+, the corresponding runtimes are 695\,s, 1{,}249\,s, and 3{,}575\,s, yielding $1.80\times$ and $5.14\times$ speedups. At this depth, \texttt{ConvRSI} generates 17.3\% fewer tokens on MATH-500 and 4.4\% fewer on MBPP+ than RecursiveMAS.

\begin{table*}[h!]
\centering
\caption{
Performance and inference efficiency across recursion depths on MATH-500 and MBPP+. All results are reproduced using NVIDIA RTX 4090D GPUs. Accuracy is reported as a percentage (\%), time in seconds (s), and token usage as generated tokens.
}
\label{tab:main_results}

\small
\setlength{\tabcolsep}{4pt}
\renewcommand{\arraystretch}{1.10}

\begin{adjustbox}{max width=\textwidth}
\begin{tabular}{lcccccc}
\toprule
\multirow{2}{*}{Method}
& \multicolumn{3}{c}{MATH-500}
& \multicolumn{3}{c}{MBPP+} \\
\cmidrule(lr){2-4}
\cmidrule(lr){5-7}
& Acc. (\%) $\uparrow$
& Time (s) $\downarrow$
& Tokens $\downarrow$
& Pass@1 (\%) $\uparrow$
& Time (s) $\downarrow$
& Tokens $\downarrow$ \\
\midrule

\multicolumn{7}{l}{\textit{Recursion round $r=1$}} \\
\midrule
Recursive-TextMAS (text-only)
& 71.4 & 2254 & 1195
& \textbf{30.4} & 1489 & 1163 \\

RecursiveMAS (latent)
& 70.1 & 1409 & 558
& 29.1 & 776 & 603 \\

\texttt{ConvRSI} (Ours)
& \textbf{71.6} & \textbf{1162} & \textbf{471}
& 30.0 & \textbf{431} & \textbf{559} \\

\midrule
\multicolumn{7}{l}{\textit{Recursion round $r=2$}} \\
\midrule
Recursive-TextMAS (text-only)
& 71.0 & 3761 & 2201
& 30.0 & 3030 & 1879 \\

RecursiveMAS (latent)
& 70.9 & 1799 & 482
& 29.5 & 982 & 566 \\

\texttt{ConvRSI} (Ours)
& \textbf{72.4} & \textbf{1255} & \textbf{464}
& \textbf{30.3} & \textbf{549} & \textbf{508} \\

\midrule
\multicolumn{7}{l}{\textit{Recursion round $r=3$}} \\
\midrule
Recursive-TextMAS (text-only)
& 69.6 & 4721 & 3088
& 29.1 & 3575 & 2548 \\

RecursiveMAS (latent)
& 71.4 & 2658 & 556
& 32.1 & 1249 & 611 \\

\texttt{ConvRSI} (Ours)
& \textbf{75.0} & \textbf{1459} & \textbf{460}
& \textbf{32.9} & \textbf{695} & \textbf{584} \\

\bottomrule
\end{tabular}
\end{adjustbox}

\vspace{3pt}

\end{table*}

\subsection{Generalization Across Collaboration Patterns}



We evaluate whether the gains from \texttt{ConvRSI} extend beyond the sequential planner--critic--solver configuration to a mixture in which a summarizer combines the outputs of code, science, and mathematics specialists. Table~\ref{tab:mas_config} lists the agents in each configuration. Within each configuration, we compare \texttt{ConvRSI} with the released RecursiveMAS checkpoints~\citep{recursivemas2026}.

\begin{table}[h!]
\centering
\small
\caption{Agent roles and model assignments for the two collaboration
patterns evaluated with \texttt{ConvRSI}. Sequential-Light uses a
planner, critic, and solver in sequence. In the mixture, a summarizer
combines outputs from code, science, and mathematics specialists.}
\label{tab:mas_config}
\resizebox{\textwidth}{!}{
\begin{tabular}{l|l|l}
\toprule
\textbf{Collaboration Pattern} & \textbf{Role} & \textbf{Model Size \& Version} \\
\midrule

\multirow{3}{*}{Sequential Style (Light)}
& Planner & Qwen3-1.7B \citep{yang2025qwen3} \\
& Critic & Llama-3.2-1B-Instruct \citep{grattafiori2024llama} \\
& Solver & Qwen2.5-Math-1.5B-Instruct \citep{qwen2.5} \\
\midrule

\multirow{4}{*}{Mixture Style}
& Code Specialist & Qwen2.5-Coder-3B-Instruct \citep{hui2024qwen2} \\
& Science Specialist & BioMistral-7B \citep{labrak2024biomistral} \\
& Math Specialist & DeepSeek-R1-Distill-Qwen-1.5B \citep{qwen2.5} \\
& Summarizer & Qwen3.5-2B \citep{yang2025qwen3} \\

\bottomrule
\end{tabular}
}
\end{table}



 

\begin{table}[h!]
\centering
\small
\setlength{\tabcolsep}{3pt}
\caption{MATH-500 accuracy and MBPP+ pass@1 (\%) by collaboration pattern. Bold marks the higher score in each column.}
\label{tab:gen_results}
\begin{adjustbox}{max width=\linewidth}
\begin{tabular}{@{}lcccc@{}}
\toprule
& \multicolumn{2}{c}{Sequential (Light)} & \multicolumn{2}{c}{Mixture} \\
\cmidrule(lr){2-3}
\cmidrule(lr){4-5}
Method & MATH-500 & MBPP+ & MATH-500 & MBPP+ \\
\midrule
RecursiveMAS            & 71.4 & 32.1 & 60.6 & 47.4 \\
\texttt{ConvRSI}        & \textbf{75.0} & \textbf{32.9} & \textbf{65.8} & \textbf{50.3} \\
\bottomrule
\end{tabular}
\end{adjustbox}
\end{table}

Table~\ref{tab:gen_results} shows that \texttt{ConvRSI} improves on RecursiveMAS under both collaboration patterns. In the sequential configuration, it gains 3.6 percentage points on MATH-500 and 0.8 points on MBPP+. The gains are larger in the mixture: MATH-500 accuracy rises from 60.6\% to 65.8\% (+5.2 points), and MBPP+ pass@1 rises from 47.4\% to 50.3\% (+2.9 points). Thus, the improvement observed with sequential collaboration also holds when a summarizer combines specialist outputs.

\subsection{Comparison with Alternative Architectures}
\label{sec:alternative_architectures}

We compare \texttt{ConvRSI} with three alternative approaches to self-improvement. ModularRSI~\citep{wu2026modularrsi} uses successful and failed execution trajectories to improve individual modules of an agent's harness, such as tool use and context management. Hermes+~\citep{xue2026pastbench} retains and updates memories and procedural skills for reuse on later tasks. MASS~\citep{lee2026mass} searches over multi-agent workflows and fine-tunes its model on trajectories generated by those workflows. We also include Recursive-TextMAS~\citep{recursivemas2026} as a text-based comparison within the Sequential-Light configuration.

\begin{table*}[h!]
\centering
\small
\caption{Comparison on mathematical reasoning and code generation, grouped by method family. The second column identifies the solver backbone or agent configuration used in our experiments. Bold marks the highest score in each benchmark column.}
\label{tab:comp_results}
\begin{tabular}{@{}llcc@{}}
\toprule
Method & Solver or agent configuration
& MATH-500 accuracy (\%) & MBPP+ pass@1 (\%) \\
\midrule
\multicolumn{4}{@{}l}{\textit{Modular harness optimization}} \\
ModularRSI & Qwen3.5-4B       & 67.1 & 32.1 \\
\midrule
\multicolumn{4}{@{}l}{\textit{Persistent memory and skills}} \\
Hermes+ & Sequential-Light solver & 64.6 & 28.6 \\
Hermes+ & Qwen3.5-4B              & 67.2 & 27.3 \\
\midrule
\multicolumn{4}{@{}l}{\textit{Multi-agent workflow optimization and self-training}} \\
MASS (Level-0) & Qwen2.5-Math-1.5B-Instruct & 62.8 & 23.5 \\
MASS (Level-1) & Qwen2.5-Math-1.5B-Instruct & 68.7 & 28.7 \\
\midrule
\multicolumn{4}{@{}l}{\textit{Sequential multi-agent collaboration}} \\
Recursive-TextMAS ($r=3$) & Sequential-Light team & 69.6 & 29.1 \\
\texttt{ConvRSI} ($r=3$)  & Sequential-Light team & \textbf{75.0} & \textbf{32.9} \\
\bottomrule
\end{tabular}
\end{table*}

Table~\ref{tab:comp_results} shows that \texttt{ConvRSI} achieves the highest scores among the configurations evaluated with Qwen solvers, reaching 75.0\% accuracy on MATH-500 and 32.9\% pass@1 on MBPP+. The closest comparison is Recursive-TextMAS: with the same Sequential-Light team and three recursion rounds, \texttt{ConvRSI} gains 5.4 percentage points on MATH-500 and 3.8 points on MBPP+. MASS also improves from Level-0 to Level-1 on both benchmarks, gaining 5.9 and 5.2 points, respectively. At Level-1, it trails \texttt{ConvRSI} by 6.3 points on MATH-500 and 4.2 points on MBPP+.


\subsection{Efficiency Analysis}
\label{sec:efficiency_analysis}

\begin{figure*}[h!]
    \centering
    \includegraphics[width=\textwidth]{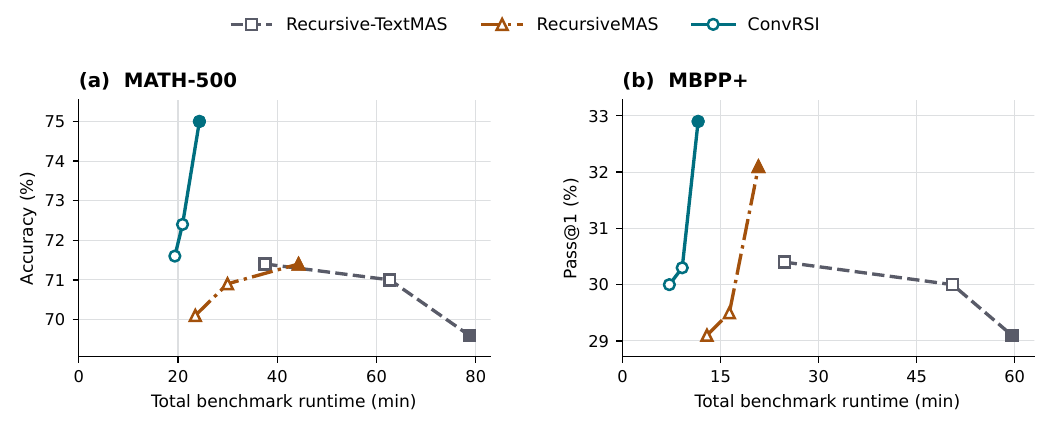}
    \caption{Performance versus total benchmark runtime. Along each line, points from left to right denote recursion depths $r=1,2,3$, with a filled marker at $r=3$. Points higher and further to the left indicate higher scores and shorter runtimes.}
    \label{fig:runtime_performance}
\end{figure*}

\paragraph{Performance at a comparable runtime.} Figure~\ref{fig:runtime_performance} shows how much additional time each method spends as recursion depth increases and how its score changes. On MATH-500, three-round \texttt{ConvRSI} reaches 75.0\% accuracy in 1{,}459\,s. This is 4.9 percentage points above one-round RecursiveMAS (70.1\%) for only 50\,s of additional runtime over the full benchmark. On MBPP+, three-round \texttt{ConvRSI} reaches 32.9\% pass@1 in 695\,s, finishing sooner than either one-round baseline: RecursiveMAS reaches 29.1\% in 776\,s and Recursive-TextMAS reaches 30.4\% in 1{,}489\,s.

\begin{figure*}[h!]
    \centering
    \includegraphics[width=\textwidth]{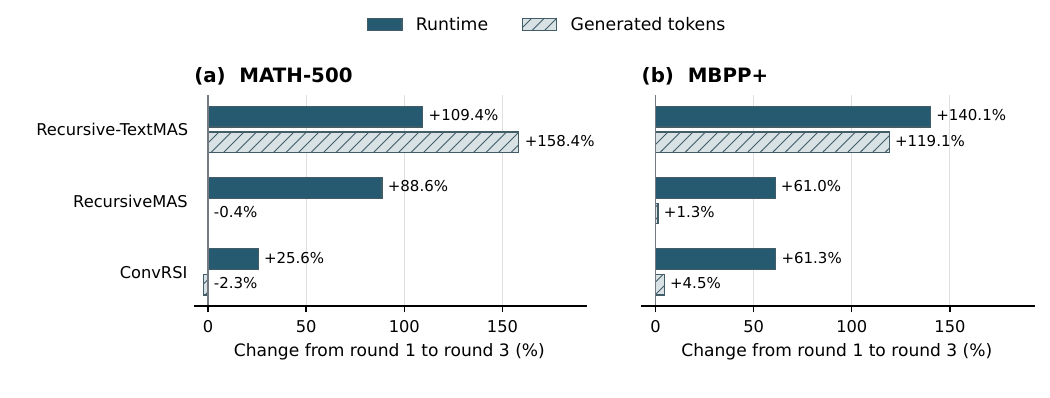}
    \caption{Change in total runtime and mean generated tokens per problem from $r=1$ to $r=3$. Each percentage is relative to the same method's one-round value on the same benchmark.}
    \label{fig:depth_cost}
\end{figure*}

\paragraph{How inference cost grows with depth.} Figure~\ref{fig:depth_cost} compares each method with its own one-round configuration. On MATH-500, increasing the depth from one to three rounds raises the runtime of \texttt{ConvRSI} by 25.6\%, compared with 88.6\% for RecursiveMAS and 109.4\% for Recursive-TextMAS. On MBPP+, the relative runtime increases for \texttt{ConvRSI} and RecursiveMAS are nearly identical (61.3\% and 61.0\%), but \texttt{ConvRSI} starts from a shorter one-round runtime and remains faster at three rounds. Generated-token counts change little for \texttt{ConvRSI} across these depths ($-2.3\%$ on MATH-500 and $+4.5\%$ on MBPP+), whereas they more than double for Recursive-TextMAS on both benchmarks.

\section{Conclusion}
\label{sec:conclusion}

We presented \texttt{ConvRSI}, a framework that restructures multi-agent self-improvement through coupled intra-agent and inter-agent convolution. By jointly refining agent features and aggregating messages across a collaboration graph, \texttt{ConvRSI} supports individual refinement and collective evolution in latent space. Its matrix-based operations expose opportunities for parallel computation while accommodating different multi-agent collaboration patterns. Experiments show that \texttt{ConvRSI} improves task performance while reducing computational costs and latency across different multi-agent collaboration patterns. Future work will explore more diverse agent collaboration patterns and extend \texttt{ConvRSI} to large-scale agent swarms.


\bibliography{ref}
\bibliographystyle{plainnat}

\appendix

\section{Author List}
\label{app:extra}

\textbf{Core Contributors.} Zheng Wang, Yuang Liu, Sachith Abeywickrama, Yangkai Ding, Tao Yu, Xinyu Zhang, Hongbo Wang, Qi Ye, Xuefeng Jin, Zhangchun Zhao

\end{document}